\documentclass[conference]{IEEEtran}
\IEEEoverridecommandlockouts
\usepackage{verbatim}
\usepackage{amsfonts}
\usepackage{amssymb}
\usepackage{stfloats}
\usepackage{cite}
\usepackage{graphicx}
\usepackage{psfrag}
\usepackage{amsmath}
\usepackage{array}
\usepackage{epstopdf}
\usepackage{authblk}
\usepackage{graphicx} 
\usepackage{amsthm} 
\usepackage{lipsum}
\usepackage{verbatim} 
\usepackage{authblk}
\usepackage{mathtools}
\usepackage{cuted}
\usepackage{booktabs}

\usepackage{amsmath}

\usepackage{array}

\usepackage{url}
\usepackage{graphicx,amsmath,amssymb,amsfonts}
\usepackage{algorithmic,algorithm}

\def\BibTeX{{\rm B\kern-.05em{\sc i\kern-.025em b}\kern-.08em
    T\kern-.1667em\lower.7ex\hbox{E}\kern-.125emX}}

\usepackage{adjustbox}
\usepackage{multirow}
\begin{document}

\title{Large AI Model-Enabled Generative Semantic Communications for Image Transmission\\

\thanks{The work of Zhijin Qin is supported by the National Key Research and Development Program of China under Grant 2023YFB2904300 and in part by the National Natural Science Foundation of China under Grant 62293484.

All authors are with the Department of Electronic Engineering, Tsinghua University, Beijing 100084, China (email: ma-qy22@mails.tsinghua.edu.cn, niwanli@tsinghua.edu.cn, qinzhijin@tsinghua.edu.cn).}}

\author{Qiyu~Ma,~\IEEEmembership{Student Member,~IEEE, }
Wanli Ni,~\IEEEmembership{ Member, IEEE} 
and~Zhijin~Qin,~\IEEEmembership{Senior~Member,~IEEE}
}
	
\maketitle

\begin{abstract}
The rapid development of generative artificial intelligence (AI) has introduced significant opportunities for enhancing the efficiency and accuracy of image transmission within semantic communication systems. Despite these advancements, existing methodologies often neglect the difference in importance of different regions of the image, potentially compromising the reconstruction quality of visually critical content. To address this issue, we introduce an innovative generative semantic communication system that refines semantic granularity by segmenting images into key and non-key regions. Key regions, which contain essential visual information, are processed using an image oriented semantic encoder, while non-key regions are efficiently compressed through an image-to-text modeling approach. Additionally, to mitigate the substantial storage and computational demands posed by large AI models, the proposed system employs a lightweight deployment strategy incorporating model quantization and low-rank adaptation fine-tuning techniques, significantly boosting resource utilization without sacrificing performance. Simulation results demonstrate that the proposed system outperforms traditional methods in terms of both semantic fidelity and visual quality, thereby affirming its effectiveness for image transmission tasks.

\end{abstract}


\section{Introduction}
The sixth-generation (6G) wireless systems necessitates high-quality multi-modal communications with ultra-high data rates, particularly for emerging applications requiring data-intensive image and video transmission \cite{dang2020should}.
However, the dramatically increasing bandwidth requirements and the computational complexity of downstream tasks pose severe challenges to traditional bit-oriented communication systems.
To address these limitations, semantic communication has re-emerged as a disruptive paradigm that fundamentally rethinks data transmission \cite{qin2024semantic}. Unlike traditional bit-oriented approaches, semantic communication optimizes information exchange by transmitting task-relevant semantic features rather than raw data, thereby enhancing spectral efficiency while maintaining reconstruction fidelity under resource constraints\cite{qin2021semantic}.
This paradigm shift towards meaning-centric transmission has gained new research momentum as a potential solution to meet the stringent requirements of 6G networks \cite{zhang2022toward}. 

Recently, generative artificial intelligence (GAI) has emerged as a transformative catalyst for advancing semantic communication, fundamentally redefining its system architecture.
Unlike traditional deep learning-based approaches focused on feature compression and channel adaptation \cite{DL-SEM}, GAI leverages its intrinsic capability to model high-dimensional data distributions, enabling holistic comprehension of latent semantic patterns rather than localized feature extraction \cite{xia2025generative}.
Specifically, GAI enables noise-resistant reconstruction, context-aware data inpainting, and adaptive regeneration \cite{zhang2024unified}.
These advantages enhance the resilience of semantic communication systems in degraded channels.
By embedding learned semantic priors directly into the communication process, GAI-driven framework surpasses traditional error correction methods and ensures human-centric semantic fidelity even under extreme channel impairments \cite{liang2024generative}. 

In the literature, research efforts have explored diverse semantic communication scenarios, such as compressed image reconstruction \cite{Generative2025Yuan} and video conferencing \cite{Multimodal2025Tong}. Generative models have demonstrated significant advantages in these areas by substantially reducing bandwidth requirements while preserving communication integrity.
The authors of \cite{Jiang2025visual} proposed a visual data transmission framework to demonstrate the potential of large generative models in improving transmission efficiency and quality. However, this framework lacked fine-grained semantic element differentiation, resulting in limited fidelity of key information and an absence of end-to-end lightweight optimization design.
In another approach, the authors of \cite{liang2025vision} leveraged vision-based GAI to enhance multiple-input multiple-output (MIMO) semantic communications.
However, their prompt-based reconstruction scheme encountered challenges in adaptability and generalization.
The authors of \cite{10966439} proposed a cross-modal panoramic transmission scheme for super-resolution video applications on mobile platforms.
Nevertheless, this work similarly failed to establish a fine-grained semantic differentiation mechanism.

Considering the computational and storage requirements of these large generative models, model lightweight is essential to achieve large-scale edge deployment of GAI-based semantic communication systems.
In this direction, researchers have made notable progress in significantly reducing computational demands through structural optimization and algorithmic innovation.
For example, the authors of \cite{zhang2024edgeshard} proposed EdgeShard, a distributed inference framework for edge computing that enables efficient large language model (LLM) execution by leveraging collaborative computation across edge devices. This approach effectively addresses resource constraints in edge environments by partitioning and distributing model workloads.
In parallel, the authors of \cite{yao2024minicpm} introduced MiniCPM-V, a compact multi-modal LLM  designed to operate at near-state-of-the-art performance levels on mobile devices. 
However, their technical designs remain largely decoupled from the semantic compression requirements inherent to communication systems.
Specifically, they lack a joint optimization mechanism that accounts for both channel characteristics and semantic fidelity, limiting their effectiveness in end-to-end semantic communication pipelines.
In addition, existing lightweight schemes focus on general-purpose computation or storage optimization, and fail to effectively consider the core requirements of cross-modal mapping and semantic entropy compression.



Motivated by the aforementioned discussions, this study aims to design a novel generative semantic communication framework for image transmission, with a focus on capturing the varying semantic importance of different image regions to enhance fine-grained semantic representation. To facilitate practical deployment on resource-constrained devices, we further develop a model lightweight strategy that significantly reduces computational overhead while preserving task performance.
The main contributions of this paper are summarized as follows:
\begin{itemize}
    \item
    We propose a generative semantic communication framework that identifies the minimal set of essential image features for transmission, while encoding the remaining residual information as structured text prompts. On the receiver side, generative models utilize these semantic anchors to reconstruct high-fidelity images. By incorporating constrained randomness into the generation process, the framework effectively bypasses classical rate-distortion limitations.
    
    \item
    To enhance deployment efficiency on edge devices, we design a three-stage optimization strategy that integrates quantized  low-rank adapters (QLoRA) and fine-tuning techniques. This strategy enables effective model compression tailored for image transmission tasks, while maintaining performance close to the original baseline.
    
    \item
    We conduct experiments to evaluate the proposed framework using various generation schemes and model configurations. The simulation results demonstrate that our framework achieves consistently superior performance across all key metrics, particularly under low SNR conditions, where it significantly outperforms conventional DeepJSCC approaches. Furthermore, ablation studies confirm the critical role of each transmitted component in achieving optimal system performance.
\end{itemize}

The remainder of this paper is organized as follows: Section~\ref{sec:GSCS} presents the architecture of our generative semantic communication framework. Section~\ref{sec:ML} describes the model lightweight and fine-tuning method. Section~\ref{sec:Simu} provides an analysis of the experimental results. Finally, Section~\ref{sec:conc} concludes this paper.

\begin{figure} [!t]
	\centering
	\includegraphics[width=3.5 in]{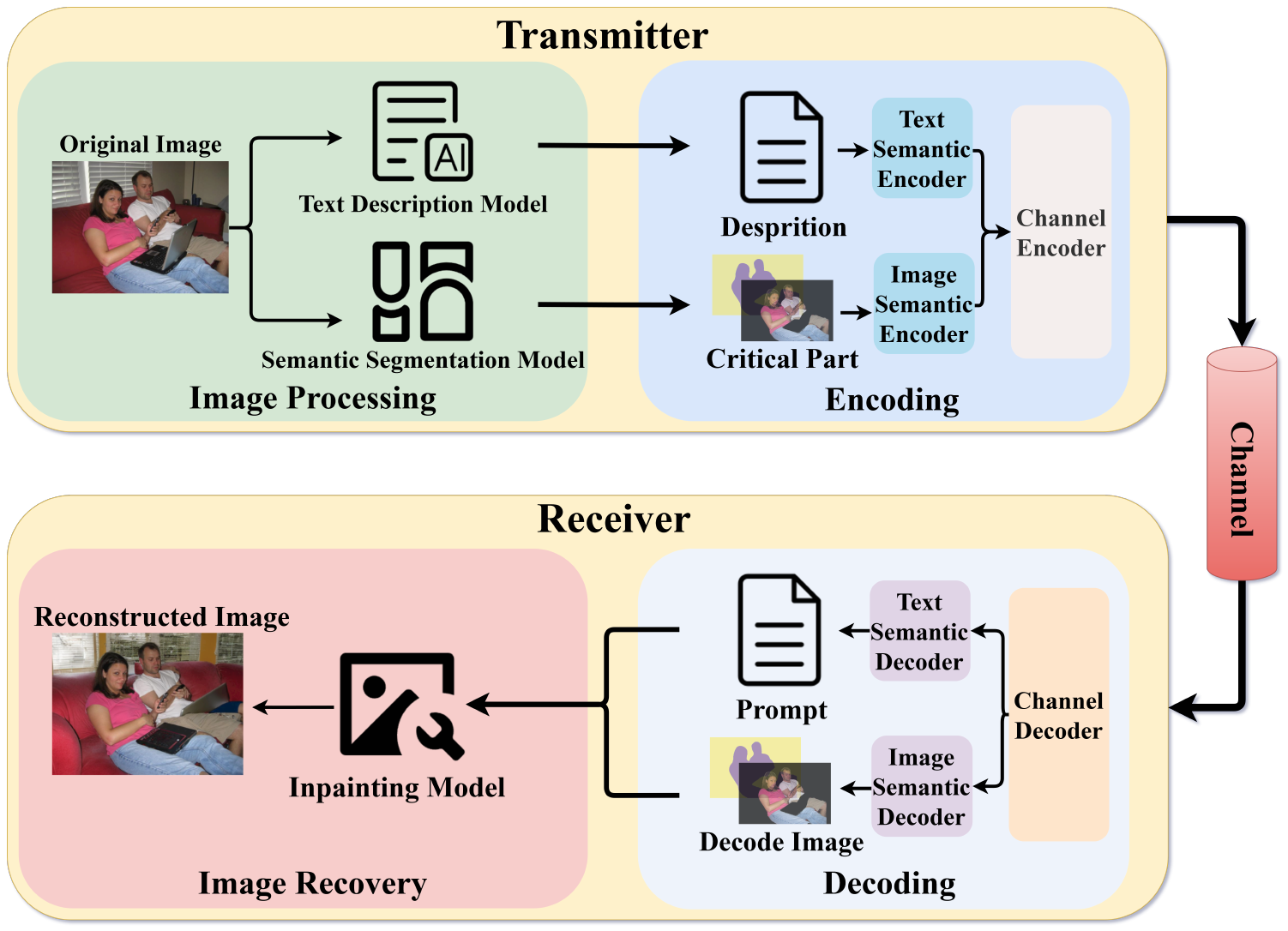}
	\caption{An illustration of the proposed generative semantic communication system for efficient image transmission over wireless networks.}
	\label{Fig.1}
\end{figure}

\section{Generative Semantic Communication System}\label{sec:GSCS}
As shown in Fig. \ref{Fig.1}, we propose a generative semantic communication system aimed at improving the efficiency of image transmission over wireless networks.
The core innovation of the proposed system lies in its semantic-aware image segmentation mechanism, which divides the input image into critical regions containing key objects and scene elements, and non-critical background regions.Specifically, key regions usually refer to the regions containing specified objects or foreground elements identified by semantic segmentation models, which carry the core semantic information of the image.
Specifically, critical regions are transmitted using image semantic encoder, non-critical parts are encoded as structured text prompts to preserve their semantic context.
These text prompts then guide the generative model at the receiver to reconstruct the high-fidelity image through the semantic-guided generation process.

Mathematically, we define the input image as $ X \in \mathbb{R}^{H \times W \times 3} $, where $ H $ and $ W $ denote its height and width, respectively.
The image follows an underlying distribution \( p_X \) and is processed through a feature segmentation module to decompose it into semantically meaningful components. Specifically, the decomposition is given by
\begin{equation}
	Y_I = \mathcal{S}_\alpha(X) \in \mathbb{R}^{m}, \quad Y_T = \mathcal{D}_\psi(X) \in \mathbb{R}^n,
\end{equation}
where \( \mathcal{S}_\alpha(\cdot) \) represents the semantic feature extraction function that preserves an \(\alpha\)-proportion of the original image's features. Here, \( m = \alpha \times \lceil H \rceil \times \lceil W \rceil \times 3 \), with \( \lceil \cdot \rceil \) denoting the ceiling function.
\( \mathcal{D}_\psi(\cdot) \) generates text descriptors parameterized by \( \psi \), which encode the remaining information in a structured textual format.
This process ensures that the critical semantic features are retained while the less important details are represented as structured text prompts.

The compressed representation is obtained using trainable encoders $\mathcal{E}_\theta(\cdot)$ and $\mathcal{E}_\phi(\cdot)$, which map the segmented components into a shared latent space.
The encoded signal is given by
\begin{equation}
	Y' = \mathcal{E}_\theta(Y_I) \oplus \mathcal{E}_\phi(Y_T) \in \mathbb{R}^k,
\end{equation}
where $ \oplus $ denotes vector concatenation, and $ k $ is the dimensionality of the encoded representation.
This encoded signal $ Y' $ is then transmitted over a noisy wireless channel, which is modeled as
\begin{equation}
	Y'' = h Y' + \mathbf{n},
\end{equation}
where $\mathbf{n} \sim \mathcal{N}(0, \sigma^2 \mathbf{I}_k)$ is the channel noise, $ \sigma^2 $ represents the noise variance and $ \mathbf{I}_k $ denotes the $ k \times k $ identity matrix.

\begin{figure*} [!t]
	\centering
	\includegraphics[width=6.5 in]{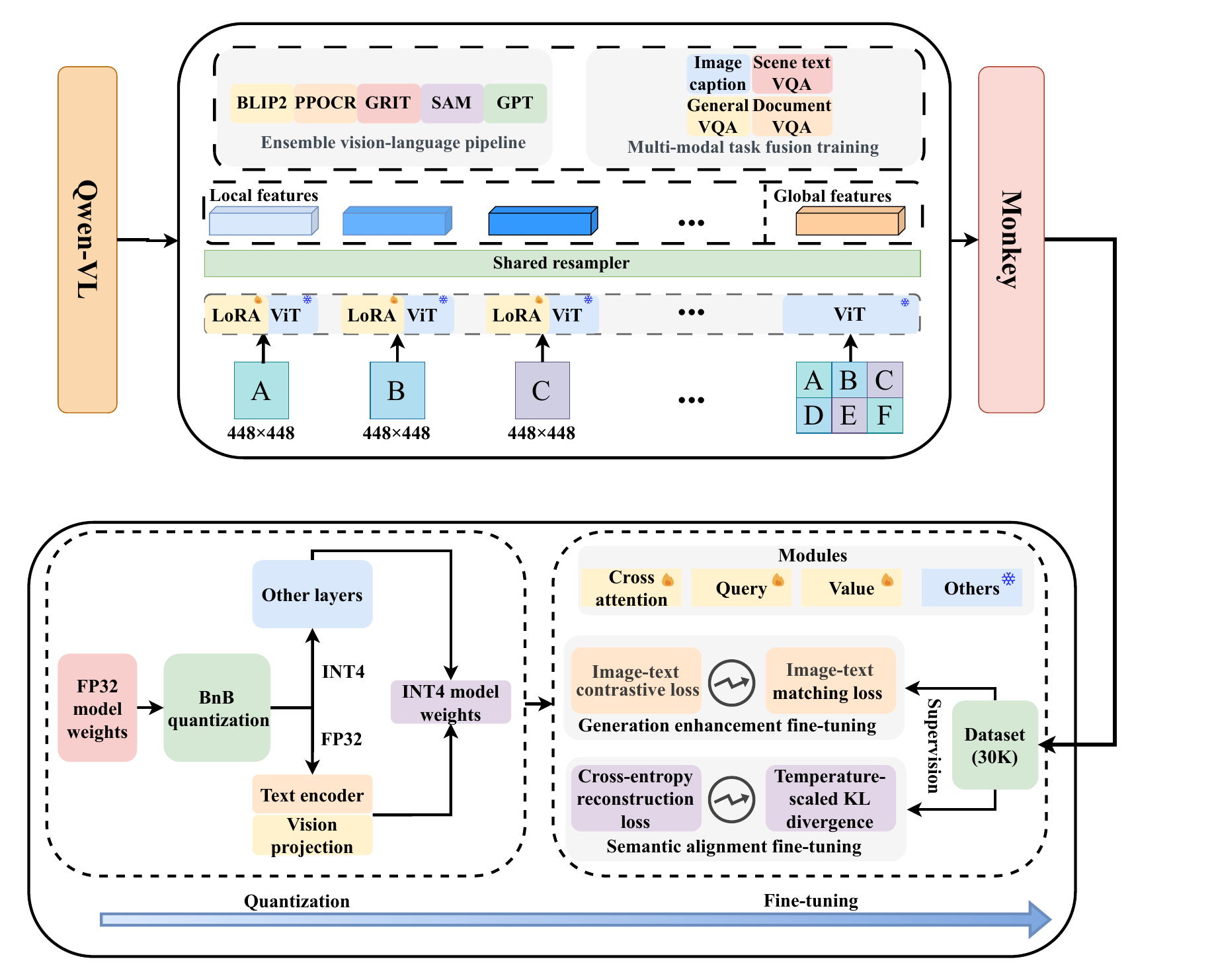}
	\caption{An illustration of the model lightweight and fine-tuning process of the proposed generative semantic communication framework. To enhance the system performance, we integrate multi-model capabilities through a multi-level description generation approach where BLIP-2 extracts global semantic information and generates preliminary descriptions, GRIT refines object attributes for local regions, PPOCR recognizes text embedded within the images, and SAM segments key regions while correlating spatial locations \cite{li2023monkey}.}
	\label{Fig.2}
	\vspace{-3 mm}
\end{figure*}

To achieve effective image reconstruction by incorporating both semantic information and controlled noise during the reverse-time diffusion process, we utilize a reverse-time stochastic differential equation (SDE). This SDE is given by
\begin{equation}
	\begin{aligned}
		dX_t &= \underbrace{\left[f_\xi(X_t, t, \hat{Y}_I, \hat{Y}_T)\right]}_{\text{Drift term}} dt + \underbrace{g(t) d\bar{w}_t}_{\text{Diffusion term}},
	\end{aligned}
\end{equation}
where \( f_\xi(\cdot) \) denotes the parameterized drift function that depends on the weights \(\xi\) and incorporates both the estimated critical image components \(\hat{Y}_I\) and the structured text prompts \(\hat{Y}_T\),
\( g(t) \) controls the noise scheduling over time,
and \( d\bar{w}_t \) represents the increment of the time-reversed Brownian motion.

The decoded features $\hat{Y}_I$ and $\hat{Y}_T$ are obtained by partitioning the noisy channel output $Y''$ as follows:
\begin{equation}
	\hat{Y}_I = \mathcal{D}_\theta(Y''_{1:k_1}) {\rm \ and \ } \hat{Y}_T = \mathcal{D}_\phi(Y''_{k_1+1:k}),
\end{equation}
where $\mathcal{D}_\theta(\cdot)$ and $\mathcal{D}_\phi(\cdot)$ denote trainable decoder functions. The final reconstructed image $\hat{X}$ is obtained by solving the reverse-time SDE to recover $X_0$, starting from an initial condition $X_T \sim \mathcal{N}(0, I)$, i.e., $\hat{X} = X_0$.
Then, the end-to-end expected distortion satisfies the following bound:
\begin{equation}
	\mathbb{E}[d(X,\hat{X})] \leq 
	\underbrace{C_1(1-\alpha)^{-1/2}}_{\text{Compression}} + 
	\underbrace{C_2\sigma\sqrt{k}}_{\text{Channel}} + 
	\underbrace{C_3\epsilon_\xi}_{\text{Diffusion}},
\end{equation}
where constants $C_1$, $C_2$ and $C_3$ characterize system performance, $\sigma$ denotes the channel noise level, and $\epsilon_\xi$ quantifies the approximation error in the diffusion process.

The above process enables our system to achieve a $(1 - \alpha)$-proportion reduction in transmission overhead through semantic prioritization. Furthermore, it establishes a principled rate-distortion trade-off via $\alpha$-parameterized semantic coding.
Unlike conventional compression techniques that focus on minimizing redundancy within fixed data representations, our approach reduces the volume of transmitted information by encoding non-critical image regions as semantic prompts rather than raw pixel values.


\section{Model Lightweight and Fine-Tuning}\label{sec:ML}
Given the substantial resource consumption of large generative models, a critical practical challenge is to reduce their resource requirements while maintaining performance.
To address this, we detail a model compression and fine-tuning scheme in this section.
As illustrated in Fig. \ref{Fig.2}, the fine-tuning process of our proposed scheme integrates quantization and LoRA techniques to achieve efficient deployment of large generative models on resource-constrained devices without significant loss in performance.

\subsection{Model Lightweight}
To significantly reduce the model’s memory footprint while preserving critical functionality, we employ a mixed-precision quantization strategy using the BitsAndBytes library. In this approach, key components of the model, such as the vision projection layer and text encoder, are maintained in full FP32 precision to safeguard essential semantic information. Meanwhile, the remaining layers are quantized to the INT4 format to reduce computational and memory overhead.
Let $\mathbf{W}_{\text{ori}}$ denote the original model weights. We partition $\mathbf{W}_{\text{ori}}$ into two subsets: $\mathbf{W}_{\text{crit}}$, representing the critical components preserved in FP32, and $\mathbf{W}_{\text{qua}}$, representing the non-critical weights subject to quantization.

Using the quantization function $\mathcal{Q}_{\text{BAB}}(\cdot)$ applied to $\mathbf{W}_{\text{qua}}$, we obtain the final quantized model weights $\mathbf{W}_{\text{BLIP-INT4}}$, defined as
\begin{equation}
	\mathbf{W}_{\text{BLIP-INT4}} = 
	\begin{cases} 
		\mathcal{Q}_{\text{BAB}}(\mathbf{W}_{\text{qua}}) & \text{quantized weights} \\
		\mathbf{W}_{\text{crit}} & \text{original precision}
	\end{cases}
\end{equation}
This mixed-precision strategy enables efficient deployment on resource-constrained devices while minimizing performance degradation due to quantization.

\subsection{Model Fine-Tuning}
Following the quantization, to enhance the capabilities of the foundation models, we perform parameter-efficient fine-tuning (PEFT) using the QLoRA method \cite{dettmers2023qlora}.
%
During the fine-tuning phase, we focus on optimizing the image description task by guiding the model to balance global semantics and detailed visual information. This is achieved by fusing global features extracted from a downscaled version of the image with aggregated local features obtained from the chunked block-wise encoding.
The fine-tuning objective is guided by an adaptive loss function designed to improve the model's ability to generate high-quality images from text prompts.
The loss integrates objectives from both knowledge distillation and sequence generation tasks.
It consists of two main components: one for aligning image and text representations $\mathcal{L}_{\text{align}}$, and another for enhancing overall performance via knowledge distillation and language modeling objectives $\mathcal{L}_{\text{enhance}}$.
These components are dynamically balanced using adaptive weighting parameters $\lambda$ and $\beta$, respectively.
The corresponding loss function is given by
\begin{equation}
	\label{eq:adaptive_loss}
	\begin{aligned}
		&\mathcal{L}_{\text{align}} = \lambda\mathcal{L}_{\text{ITC}} + (1 - \lambda)\mathcal{L}_{\text{ITM}}, & \lambda \in [0,1], \\
		&\mathcal{L}_{\text{enhance}} = \beta\mathcal{L}_{\text{KL}} + (1 - \beta)\mathcal{L}_{\text{CE}}, & \beta \in [0,1],
	\end{aligned}
\end{equation}
where
$\mathcal{L}_{\text{ITC}}$ represents the image-text contrastive loss,
$\mathcal{L}_{\text{ITM}}$ is the image-text matching loss,
$\mathcal{L}_{\text{KL}}$ is the KL-divergence loss,
$\mathcal{L}_{\text{CE}}$ is the cross-entropy loss.
This adaptive loss framework ensures a balanced optimization process, facilitating strong performance in both semantic alignment and generative fidelity.

\begin{table}[t]
	\centering
	\caption{Comparison of Different Models}
	\begin{tabular}{lccc}
		\hline
		\textbf{Models} & \textbf{Parameters (M)} & \textbf{Memory (GB)}   \\
		\hline
		Monkey & 9708.05 & 18.09 \\
		BLIP & 247.41 & 0.92 \\
		Proposed & \textbf{204.95} & \textbf{0.35} \\
		\hline
	\end{tabular}
	\label{tab:model-comparison}
    \vspace{-2mm}
\end{table}

The comparison in Table \ref{tab:model-comparison} illustrates the significant performance gains achieved by the proposed scheme in terms of resource utilization and task performance. Specifically, the proposed scheme drastically reduces memory requirements, achieving a 62\% decrease compared to the original BLIP model and a 98\% reduction compared to the Monkey model. These substantial improvements in memory efficiency are accomplished without compromising on task performance.

\begin{table}[!t]
	\centering
	\caption{Parameter Setup and Evaluation Metrics}
	\label{tab:params_metrics}
	\begin{adjustbox}{max width=3.5 in}
		\begin{tabular}{@{}lll@{}}
			\toprule
			\textbf{Category} & \textbf{Metric} & \textbf{Implementation} \\
			\midrule
			\multirow{3}{*}{Model parameters} 
			& num\_beams & 3 \\
			
			& length\_penalty & 0.5 \\
			& num\_return\_sequences & 3 \\
			\midrule
			\multirow{12}{*}{Metrics}
			& Fréchet inception distance (FID) & \multirow{5}{*}{PyTorch} \\
			& Perceptual loss &  \\
			& Style loss &  \\
			& Content loss &  \\
            & Visual saliency-induced index (VSI) & \\
            \cmidrule(lr){2-3}
			& Learned perceptual similarity (LPIPS) & lpips \\
			\cmidrule(lr){2-3}
			& Structural similarity (SSIM) & scikit-image \\\cmidrule(lr){2-3}
			& Mean squared error (MSE) & \multirow{2}{*}{NumPy} \\
			& Intersection over union (IoU) &  \\\cmidrule(lr){2-3}
			& Color histogram similarity & \multirow{3}{*}{OpenCV} \\
			& Keypoint matching similarity &  \\
            & Peak signal-to-noise ratio (PSNR) & \\
			\bottomrule
		\end{tabular}
	\end{adjustbox}
    \vspace{-3mm}
\end{table}

\section{Simulation Results}\label{sec:Simu}
To evaluate the performance of our proposed generative semantic communication system in image transmission, a series of experiments are conducted. At the heart of our transmitter lies a two-step process: initially, a semantic segmentation model identifies and prioritizes critical regions within an image; subsequently, an image captioning model generates descriptive text for these regions. Upon reception, advanced generative AI models are employed to reconstruct the original images based on the received semantic information. 
For testing, we randomly selected 900 images from the PIPA dataset and supplemented them with Kodak24. In these images, foregrounds are marked as critical, simulating scenarios where certain objects or subjects hold greater significance. To thoroughly assess the effectiveness of the proposed scheme, we compare it against several baselines, including
\begin{itemize}
\item \textbf{Baseline 1 (DirectlyGen)}: Generates images using only text prompts from image captioning models.
\item \textbf{Baseline 2 (AllEdge)}: Combines text prompts with semantic segmentation masks and region labels for image generation.
\item \textbf{Baseline 3 (HumanOnly)}: Focuses exclusively on human-centric regions for generation, excluding contextual information.
\item \textbf{Conventional BPG+LDPC}: Image compression using BPG codec with LDPC channel coding for error-resilient transmission.
\item \textbf{Conventional DeepJSCC}: End-to-end neural codec implementing joint source-channel coding via deep learning\cite{xu2023deep}.
\end{itemize}

\begin{figure}[!t]
	\centering
	\includegraphics[width=3.5 in]{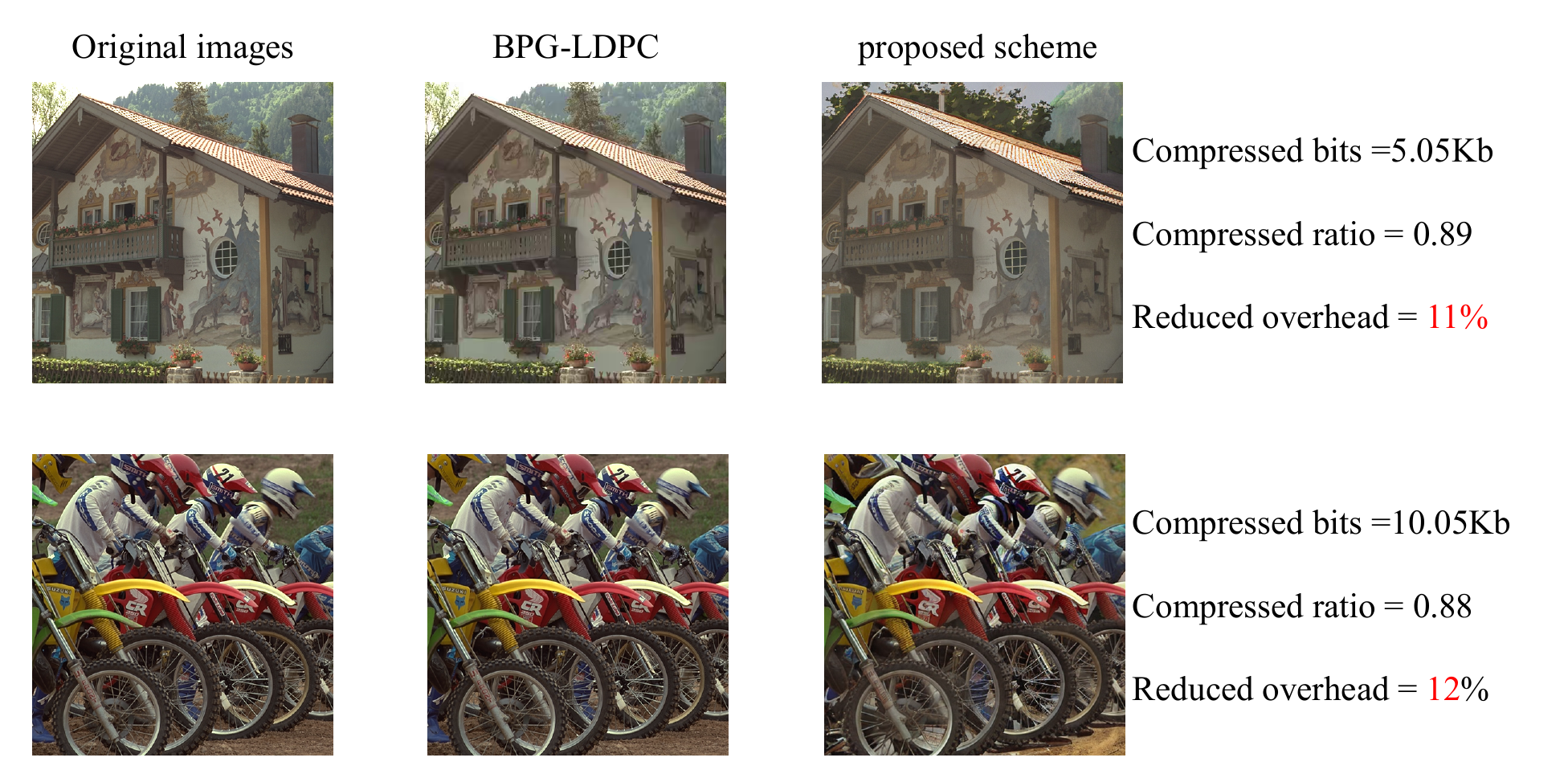}
	\caption{Simulation results of the BPG-LDPC and the proposed scheme.}
	\label{Fig3}
\end{figure}

\begin{figure}[!t]
	\centering
	\includegraphics[width=3.5 in]{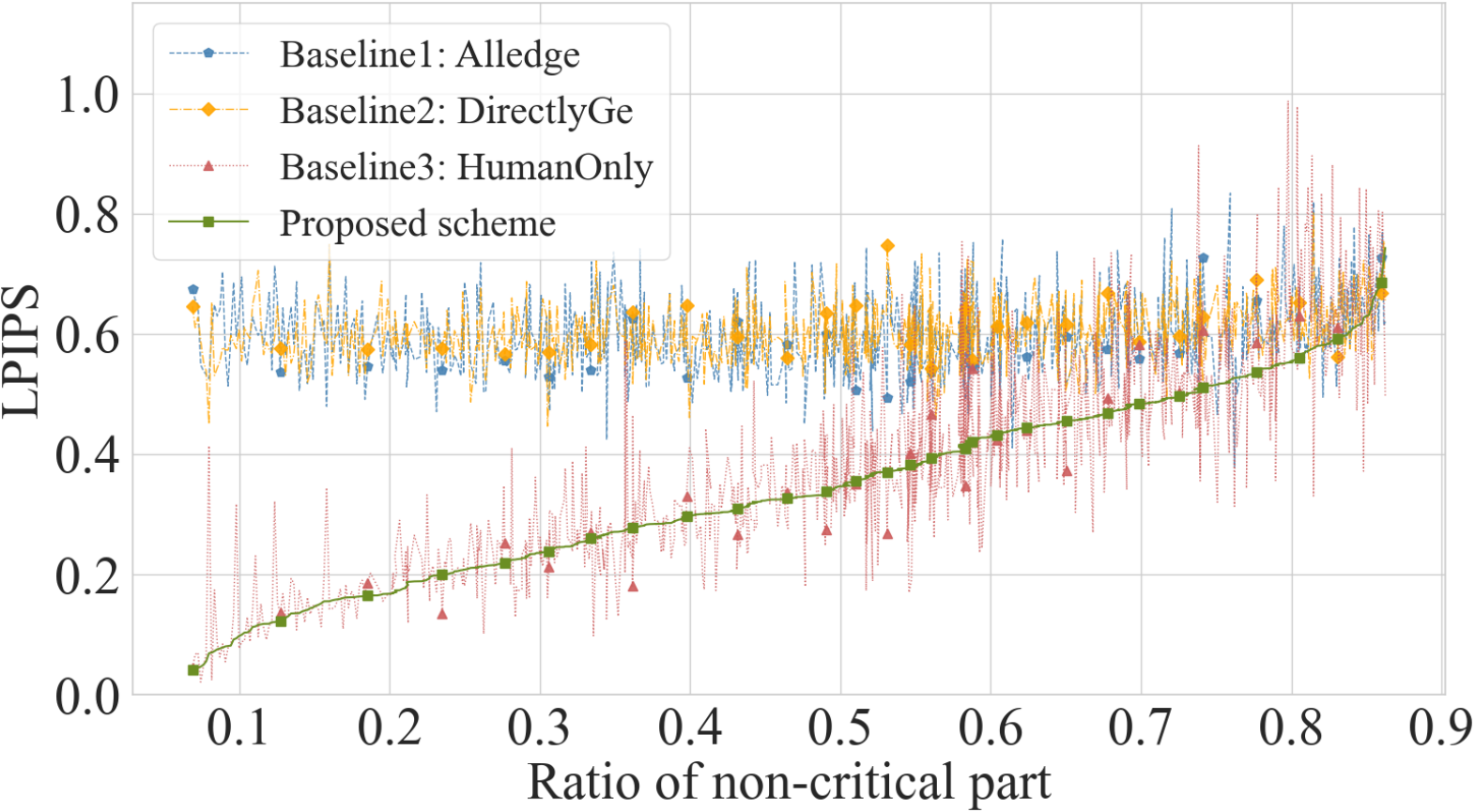}
	\caption{Performance comparison in terms of LPIPS.}
	\label{Fig4} 
	\vspace{-3 mm}     
\end{figure}
The reconstruction quality is evaluated using metrics including LPIPS ($\downarrow$) for initial assessment, and extended analysis with FID ($\downarrow$), IoU ($\uparrow$), MSE ($\downarrow$), SSIM ($\uparrow$), PSNR ($\uparrow$), VSI ($\uparrow$), keypoint similarity ($\uparrow$), perceptual/style/content loss ($\downarrow$), and color histogram distance. Metrics marked $\downarrow$ improve with lower values, while $\uparrow$ indicates better performance at higher values. Evaluation parameters are detailed in Table \ref{tab:params_metrics}.

\begin{figure} [!t]
	\setlength{\abovecaptionskip}{-0.2cm} 
	\centering
	\includegraphics[width=3.5 in]{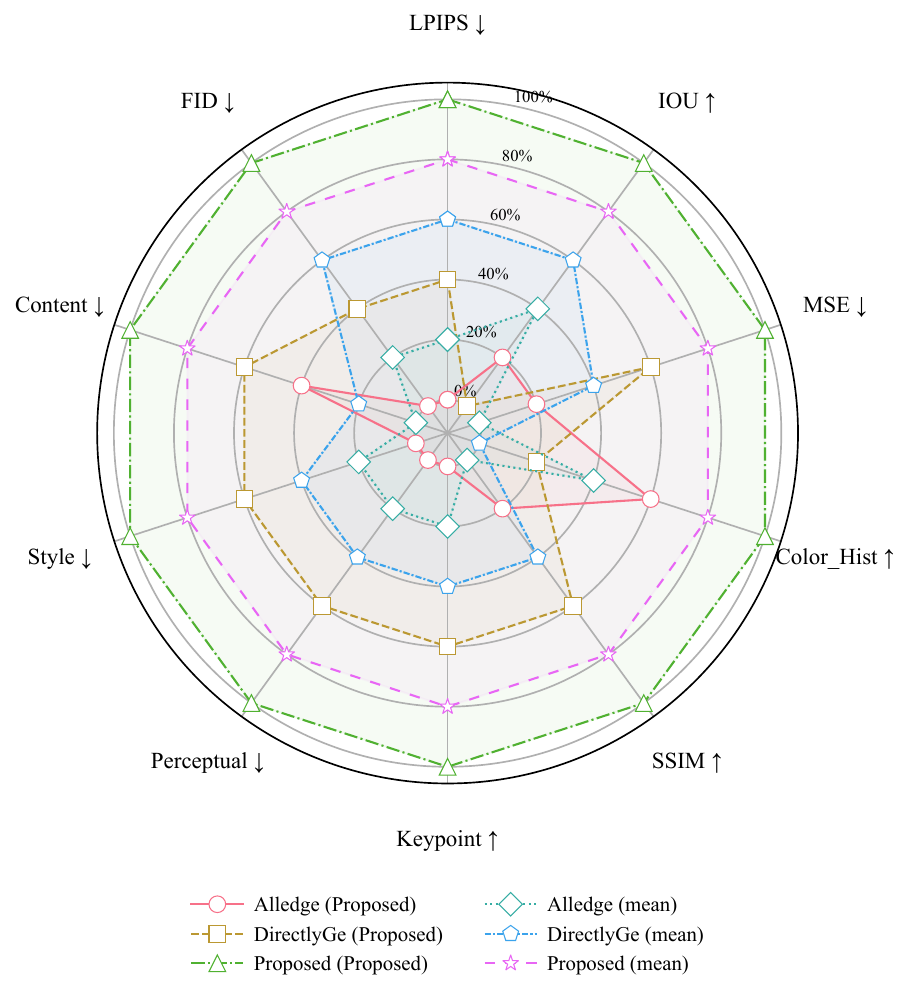}
	\caption{Multi-criteria performance comparison: A radar chart illustrating the comprehensive evaluation of various semantic communication schemes across multiple metrics. }
	\label{Qiyu_5}
	\vspace{-3 mm} 
\end{figure}

In Fig. \ref{Fig3}, we present a visual comparison between images transmitted by the BPG-LDPC methods and our proposed generative semantic communication system. For instance, when transmitting the first image, our system achieves a 11\% reduction in communication overhead, while for the second image, it reduces the overhead by 12\%.
These significant reductions are achieved without compromising the quality of the reconstructed images, as evidenced by their close resemblance to the original images.
In Fig. \ref{Fig4}, we further analyze the performance of different semantic communication schemes by plotting the ratio of non-critical parts (1-$\alpha$) against the LPIPS score. Notably, when prompts are excluded from the model input, the LPIPS curve exhibits greater variability and unpredictability, leading to less consistent image quality and fidelity. This observation highlights the critical role of incorporating prompts or additional context information into the image generation process.
By doing so, our proposed scheme achieves more stable and reliable outcomes, as reflected by the smoother and lower LPIPS curve compared to the baselines.

In Fig. \ref{Qiyu_5}, a multi-criteria radar analysis is presented. This visualization demonstrates that the proposed scheme attains the highest rank (i.e., Rank 1) across all evaluated metrics. The radar chart effectively highlights the superior performance of our approach compared to other methods, such as Alledge and DirectlyGe, reinforcing its robustness and effectiveness in image transmission tasks.

\begin{figure} [!t]
	\setlength{\abovecaptionskip}{-0.2cm} 
	\centering
	\includegraphics[width=3.5 in]{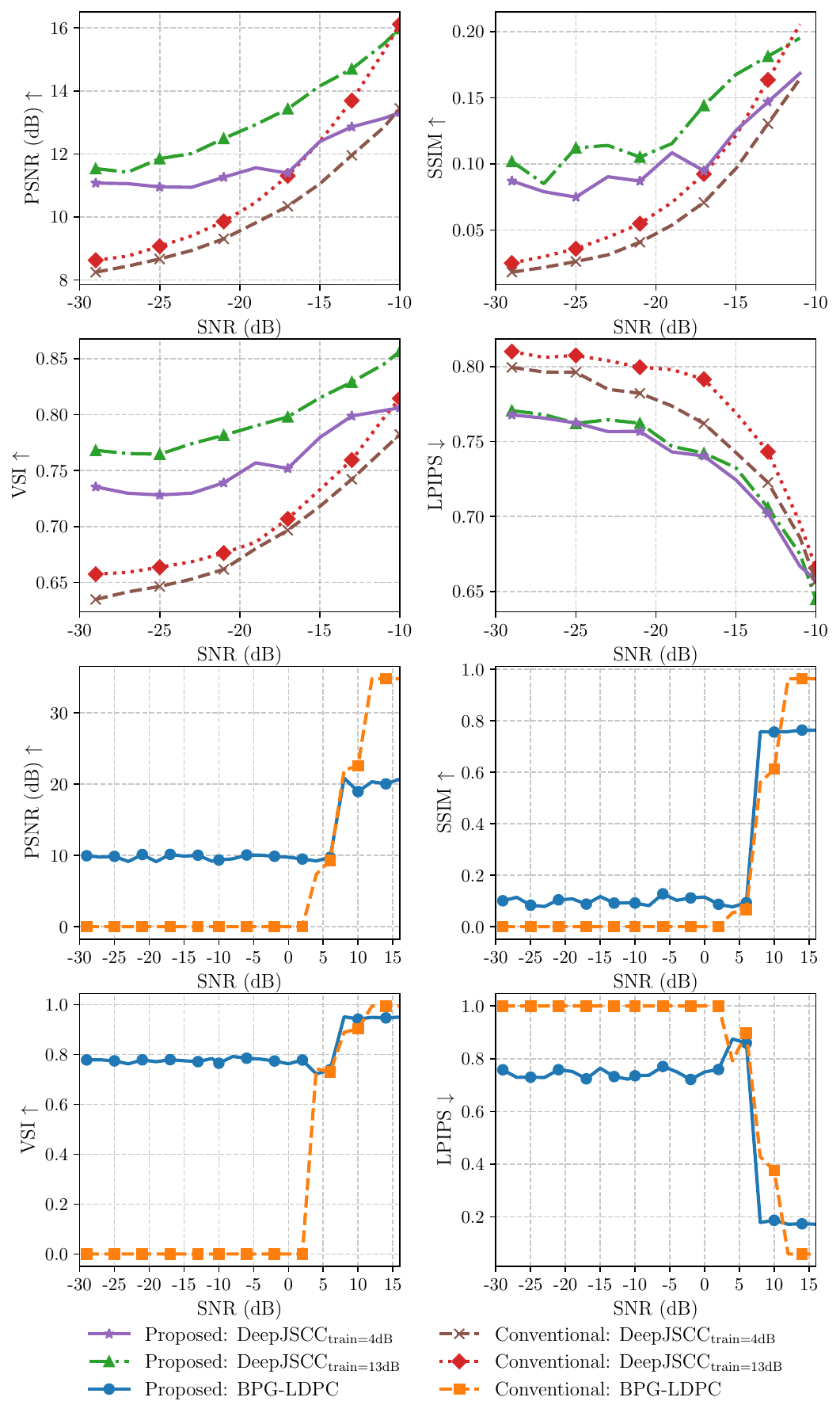}
	\caption{Performance comparison in terms of metrics against SNR}
	\label{Qiyu_6}
	\vspace{-3 mm} 
\end{figure}

%

The comparison in Fig. \ref{Qiyu_6} vividly highlights the advantages of our proposed scheme over conventional DeepJSCC and BPG-LDPC approaches under challenging channel conditions. The visualization demonstrates that our method achieves relatively improved PSNR, SSIM, and VSI scores along with reduced LPIPS values in low SNR environments, indicating superior reconstruction accuracy, enhanced structural preservation, better visual fidelity, and minimized perceptual discrepancies. These comprehensive improvements underscore the scheme's strengthened capability to maintain relatively reliable image transmission quality even under severe signal degradation scenarios, showcasing its practical effectiveness in real-world communication systems with harsh channel conditions.

\section{Conclusions}\label{sec:conc}
In this paper, we proposed a generative semantic communication system for efficient image transmission. The core idea involved decomposing images based on semantic significance, where critical regions were encoded using traditional image processing methods, while non-critical components were represented as structured text prompts. These prompts effectively guided generative AI models at the receiver to reconstruct high-fidelity images, thereby surpassing the classical rate-distortion limits. To facilitate deployment on mobile and edge devices, we introduced a model quantization and fine-tuning scheme which achieved notable improvements in computational efficiency without compromising image reconstruction accuracy, making the proposed system well-suited for resource-constrained environments.

\end{document}